\documentclass{article}
\usepackage{ijcai22}

\usepackage{times}
\usepackage{soul}
\usepackage{url}
\usepackage[hidelinks]{hyperref}
\usepackage[utf8]{inputenc}
\usepackage[small]{caption}
\usepackage{graphicx}
\usepackage{amsmath}
\usepackage{amsthm}
\usepackage{amsfonts}
\usepackage{booktabs}
\usepackage{algorithm}
\usepackage{algorithmic}
\usepackage{booktabs}
\usepackage{multirow}
\title{Neutral Utterances are Also Causes: Enhancing Conversational Causal Emotion Entailment with Social Commonsense Knowledge}

\author{
Jiangnan Li$^{1,2,}$\thanks{Joint work with Pattern Recognition Center, WeChat AI, Tencent Inc, China. }
\and
Fandong Meng$^3$\and
Zheng Lin$^{1,2,}$\thanks{Zheng Lin and Peng Fu are the corresponding authors. }\and
Rui Liu$^{1,2}$\and
Peng Fu$^{1,\dag}$\and
Yanan Cao$^{1,2}$\and
Weiping Wang$^{1}$\And
Jie Zhou$^{3}$
\affiliations
$^1$Institute of Information Engineering, Chinese Academy of Sciences, Beijing, China\\
$^2$School of Cyber Security, University of Chinese Academy of Sciences, Beijing, China\\
$^3$Pattern Recognition Center, WeChat AI, Tencent Inc, China\\
\emails
\{lijiangnan,linzheng,liurui1995,fupeng,wangweiping\}@iie.ac.cn, \{fandongmeng,withtomzhou\}@tencent.com
}

\begin{document}

\maketitle

\begin{abstract}
  \textit{Conversational Causal Emotion Entailment} aims to detect causal utterances for a non-neutral targeted utterance from a conversation. In this work, we build conversations as graphs to overcome implicit contextual modelling of the original entailment style. Following the previous work, we further introduce the emotion information into graphs. Emotion information can markedly promote the detection of causal utterances whose emotion is the same as the targeted utterance. However, it is still hard to detect causal utterances with different emotions, especially neutral ones. The reason is that models are limited in reasoning causal clues and passing them between utterances. To alleviate this problem, we introduce social commonsense knowledge (CSK) and propose a Knowledge Enhanced Conversation graph (KEC). KEC propagates the CSK between two utterances. As not all CSK is emotionally suitable for utterances, we therefore propose a sentiment-realized knowledge selecting strategy to filter CSK. To process KEC, we further construct the Knowledge Enhanced Directed Acyclic Graph networks. Experimental results show that our method outperforms baselines and infers more causes with different emotions from the targeted utterance.\footnote{ 
 The code and Appendix are available at \url{https://github.com/LeqsNaN/KEC}}
\end{abstract}
\section{Introduction}

Conversation is the most omnipresent medium for people to express daily emotions. Capturing emotions is critical for machines to understand conversations, which leads to researchers \cite{DialogueRNN,DialogueGCN,DAG} devoting into emotion recognition in conversations (ERC). However, further reasoning why an emotion is aroused in conversations is still under explored \cite{RECCON}. Understanding emotion causes is also a critical step for machines to understand conversations, which can be further utilized in other high-level tasks like empathetic generation \cite{CauseGen}. Therefore, Poria et al.~\shortcite{RECCON} propose a new task called \textbf{R}ecognizing \textbf{E}motion \textbf{C}ause in \textbf{Con}versations (RECCON). RECCON involves an important subtask called \textit{Conversational Causal Emotion Entailment} (C$_2$E$_2$). C$_2$E$_2$ detects causal utterances for a targeted utterance from conversations where the emotion is given. 

\begin{figure}
    \centering
    \scalebox{0.46}{\includegraphics{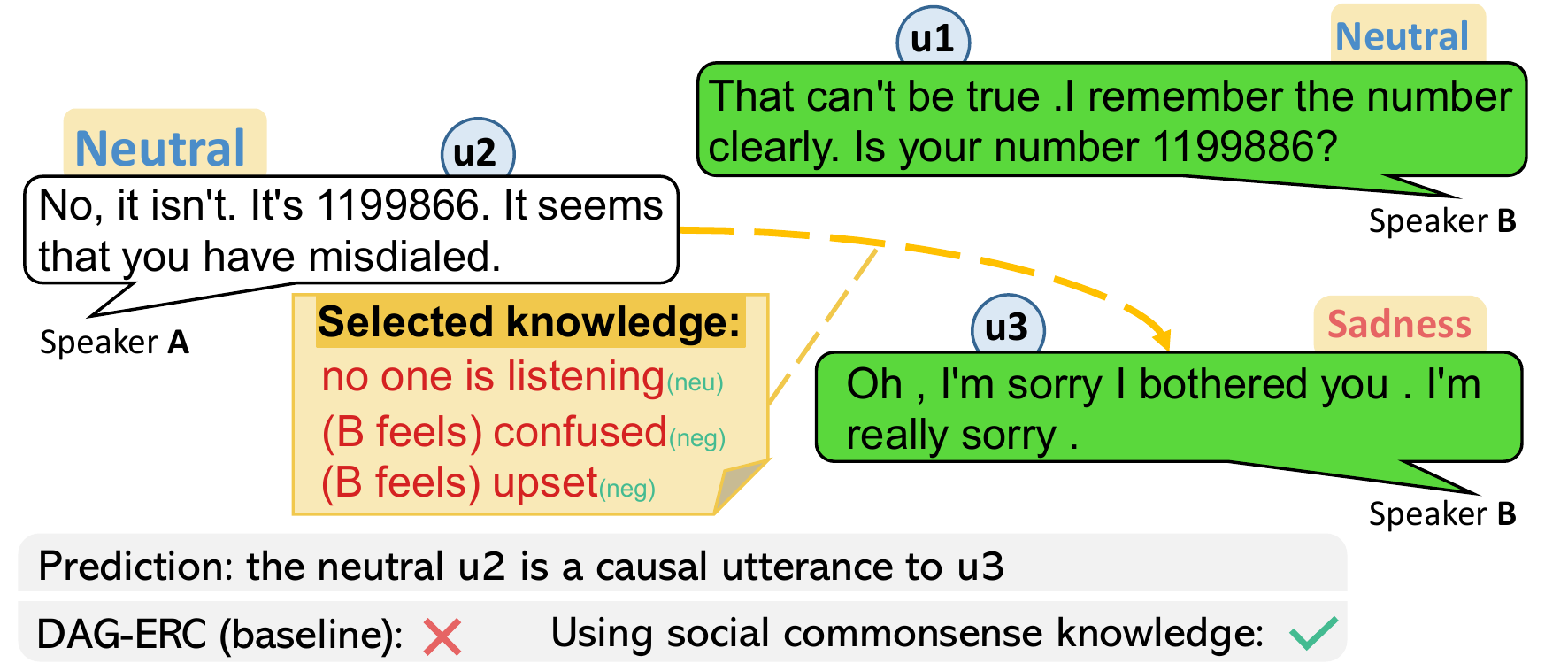}}
    \caption{A case that the baseline DAG-ERC fails while the selected commonsense knowledge can help to make a right prediction. }
    \label{case}
    \vspace{-0.2cm}
\end{figure}

Implied by the name, C$_2$E$_2$ is originally modeled as a ``text pair'' task by Poria et al.~\shortcite{RECCON}. In this way, a candidate utterance and the targeted utterance are paired plus their conversational context and the emotion. They are concatenated and then processed by a pretrained model. However, this entailment style ignores the explicit interactions between utterances, thus contextual information from conversations not fully captured. To explicitly model the interactions between utterance representations, works \cite{DialogueGCN,DAG} in ERC adopt various types of graphs for conversations, which can achieve decent performance. 

As emotion plays a crucial role in C$_2$E$_2$, it is utilized in the entailment \cite{RECCON}. When conversations are constructed as graphs, it is also critical to consider emotion information. An intuitive way is to represent emotion as emotion embedding \cite{Liang} and concatenate it with the node representation. Experiments show that emotion information can remarkably boost the performance. Specifically, we find that it can dramatically promote the detection of causal utterances whose emotion is the same as the targeted utterance. However, causal utterances with different emotions, especially neutral ones (neutral causal utterances occupy 87$\%$ of this kind of causes), is still hard to detect even with emotion information. We think the reason lies in that models are limited in reasoning causal clues and passing them between utterances. For example, in Fig.~\ref{case}, the 2nd utterance $u_{2}$ is a cause to the 3rd utterance $u_{3}$. Models cannot reason why a neutral utterance about misdialing can evoke sadness to another utterance, leading to a wrong prediction. 

To boost models' causal reasoning ability, we introduce social commonsense knowledge (CSK) \cite{ATOMIC2020} and propose a graph-based structure to properly utilize CSK. Therefore, causal utterances with different emotions from the targeted utterance can be further detected. Specifically, social CSK can infer a person's social interactions like mental states and reactions. Back to Fig.~\ref{case}, equipped with the knowledge, models can infer that $u_2$ makes the speaker of $u_3$ feel upset and then give a right prediction. To properly utilize knowledge, we propose a Knowledge Enhanced Conversation graph (KEC). KEC passes knowledge through edges. As the knowledge should align with speaker interactions and emotions, we therefore propose a speaker and sentiment realized knowledge selecting strategy, which can pick up suitable knowledge for utterances. Furthermore, to process the KEC, we extend the Directed Acyclic Graph networks (DAG) for ERC \cite{DAG} to build Knowledge Enhanced DAG networks. To verify our method, we conduct experiments on the RECCON dataset. Our method significantly outperforms the baseline of RECCON and other highly related methods. In addition, our method promotes the detection of causal utterances with different emotions from the targeted utterance. 


\begin{figure*}
    \centering
    \scalebox{0.465}{\includegraphics{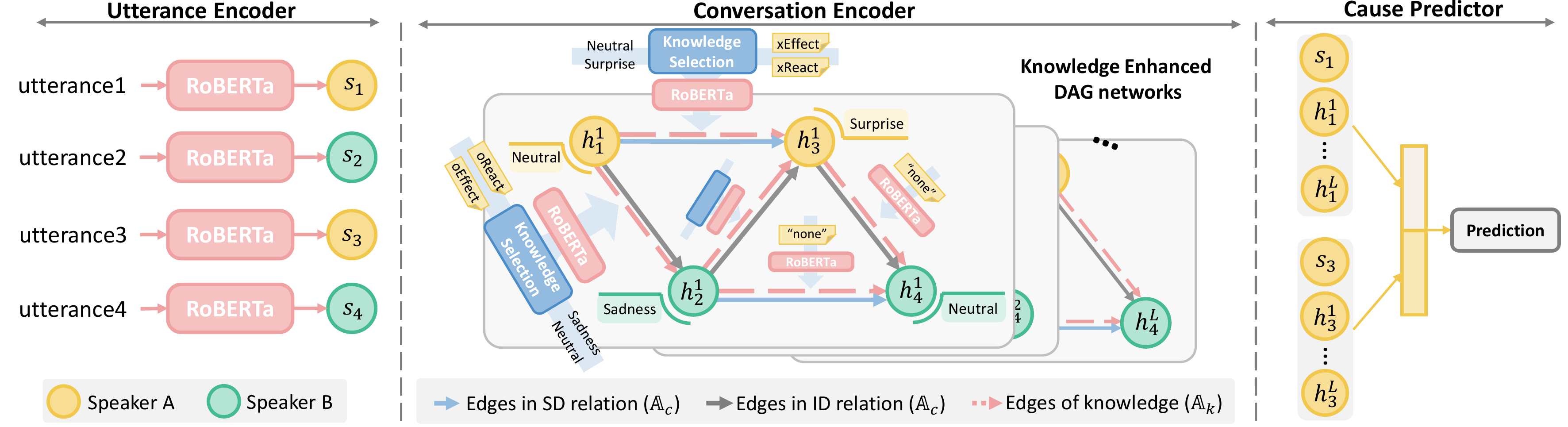}}
    \caption{The structure of our model. It contains 3 modules: (1) Utterance Encoder encodes every utterance; (2) KEC graph is constructed from a conversation and knowledge attached in KEC graph is picked up by the knowledge selecting strategy. Conversation Encoder then uses Knowledge Enhanced DAG networks to process KEC graph; (3) Cause Predictor pairs every two utterances to make predictions. }
    \vspace{-0.2cm}
    \label{model}
\end{figure*}

\section{Related Work}

\textbf{Conversational Emotion Recognition} (ERC) is a highly related task to C$_2$E$_2$. ERC tends to predict the emotion of utterances in conversations. In ERC, a main stream of works \cite{DialogueRNN,DialogueGCN,DialogXL} focuses on studying the speaker modelling. Although ERC is a hot-spot task for researchers to study conversations, further reasoning about emotions and understanding the causes in conversations is still lack of exploration. 


\textbf{Emotion Cause Extraction} (ECE) is to extract the causal clause for an emotion clause from a document and Gui et al.~\shortcite{ECE} construct the most popular ECE dataset. Ding et al.~\shortcite{PAE_DGL}; Xia et al.~\shortcite{RTHN} introduce position embeddings to model causal clauses appeared near the emotion clause. As emotion and cause are highly related, Turcan et al.~\shortcite{AdaptedKnow} propose multi-task learning frameworks for emotion recognition and cause extraction. To simultaneously extract emotion and cause clauses from documents, Xia and Ding~\shortcite{ECPE} propose a new task called Emotion Cause Pair Extraction (ECPE) and handle it in a two-stage style. To overcome the problems brought by two-stage processing, some works \cite{ECPE_2D,Rank_Emotion,SlidingWindow} propose their novel joint frameworks. Both ECPE and ECE are only extract causes in articles, which does not extend to the scene of conversations. Conversations are more complicated due to the flowing emotion dynamics and speaker participation \cite{DialogueGCN}. To study the causal factor of emotion in conversations, Poria et al.~\shortcite{RECCON} propose Recognizing Emotion Causes in Conversations (RECCON). RECCON involves two subtasks: \textit{Conversational Causal Span Extraction} (C$_2$SE) and \textit{Conversational Causal Emotion Entailment} (C$_2$E$_2$). C$_2$SE focuses on extracting causal spans from conversations, which is modelled as a machine reading comprehension task by Poria et al.~\shortcite{RECCON}. In this work, we focus on C$_2$E$_2$, the classification version of RECCON. 

\textbf{Commonsense Knowledge Utilization} (CSK) can bring rational external knowledge for models. In ERC, some works \cite{KET,COMSIC,SKAIG} study how to use CSK to help understand conversation context and mental states of speakers. In ECE, Gui et al.~\shortcite{KAG} propose Knowledge-Aware Graph using CSK to alleviate position bias brought by position embeddings. Turcan et al.~\shortcite{AdaptedKnow} use CSK to enhance ECE and Emotion Detection. In the task of Sentence Ordering, sequential commonsense knowledge is structurally utilized by Ghosal et al., \shortcite{StaCK}. However, for C$_2$E$_2$, it is still under explored how to properly use CSK. 

\section{Method}

\subsection{Task Definition}

C$_2$E$_2$ is to find causal utterances for a targeted utterance from the historical context. Given a conversation $C=[u_1,...,u_N]$, with the corresponding emotion sequence $E=[e_1,...,e_N]$ and speaker sequence  $P=[p_1,...,p_N]$, every utterance $u_i$ is paired with its contextual utterances $u_j$ ($j\leq i$). If $u_i$ is a non-neutral utterance and $u_j$ is its causal utterance, the pair $(u_i, u_j)$ is labeled with 1. Otherwise, $(u_i, u_j)$ is labeled with 0. In our setting, neutral utterances are involved as targeted utterances, leading to more negative non-causal pairs and making the task more challenging. 

\subsection{Utterance Encoder}

Similar as other conversation-related tasks (e.g., ERC), utterances in a conversation are first encoded by an utterance encoder to form utterance representations.  We utilize the pretrained model RoBERTa \cite{roberta} to encode an utterance $u_i=[w_1,...,w_{L_u}]$ with $L_u$ words. We then obtain the utterance representation by:
\begin{align}
    s_i=\mathrm{Linear}(\mathrm{Maxpooling}(\mathrm{RoBERTa}(u_i))),
\end{align}
where $s_i\in \mathbb{R}^{d_{u}}$ and $d_{u}$ is the dimension of utterance representation. 

\subsection{Conversation Encoder}

As only reviewing conversation history for utterances, the conversation can be modelled as a directed acyclic graph when not considering self-loop edges. Based on the Directed Acylic Graph (DAG) networks \cite{DAG}, we introduce our knowledge enhanced conversation graph, which is illustrated in Fig.~\ref{model}. 

\subsubsection{Knowledge Enhanced Conversation Graph}

Knowledge Enhanced Conversation graph (KEC) is formulated as $\mathcal{G}=(\mathbb{V}, \mathrm{\mathbb{A}_c}, \mathrm{\mathbb{A}_k})$. $\mathbb{V}$ denotes the set of utterance nodes, $\mathrm{\mathbb{A}_c}$ denotes the adjacency matrix of conventional utterance interactions, and $\mathrm{\mathbb{A}_k}$ denotes the adjacency matrix of knowledge passing between utterances. 

\paragraph{Utterance nodes.} $\mathbb{V}$ models all utterances as nodes $\mathrm{v}_i$ in a conversation. $\mathrm{v}_i$ contains the attribute {\tt rep} to store the utterance representation (i.e., $\mathrm{v}_i.${\tt rep}$=s_i$). 

\paragraph{Utterance interaction adjacency matrix.} Conversational context modeling is crucial to conversation-related tasks, which can enrich targeted utterances with contextual messages passed from other utterances. As only the dialogue history is considered, we construct $\mathrm{\mathbb{A}_c}$ as a lower triangle matrix. $\mathrm{\mathbb{A}_c}$ contains two attributes: {\tt item} and {\tt rel}, where {\tt item} stores 0 or 1 to denote the existence of an edge and {\tt rel} stores the relation type of an edge. 

For nodes $\mathrm{v}_i$, $\mathrm{v}_j$, $\mathrm{\mathbb{A}_c}[i, j].\texttt{item}=1$ is present if there is an edge from $\mathrm{v}_j$ to $\mathrm{v}_i$, otherwise $\mathrm{\mathbb{A}_c}[i, j].\texttt{item}=0$. For a targeted utterance, it has been studied that local contextual information is more important than that in the remote context \cite{DAG}. Therefore, a context window with size of $w_c$ is sliding through conversations. In the window, every contextual utterance sends an edge to the targeted one. 
Speaker information is another crucial factor. Speaker interactions can be modelled as two relations: self-dependency (SD) and inter-speaker dependency (ID) \cite{DialogueGCN}. For $\mathrm{v}_i$, $\mathrm{v}_j$, $\mathrm{\mathbb{A}_c}[i, j].\texttt{rel}=\mathrm{SD}$ means that the speaker identity $p_i$ is the same as $p_j$, otherwise $\mathrm{\mathbb{A}_c}[i, j].\texttt{rel}=\mathrm{ID}$. 

\paragraph{Knowledge passing adjacency matrix.} We pass the knowledge through edges to enhance the ability of inferring causes. Similar as $\mathrm{\mathbb{A}_c}$, $\mathrm{\mathbb{A}_k}$ is also a lower triangle matrix. $\mathrm{\mathbb{A}_k}$ contains two attributes: {\tt item} and {\tt klg}, where {\tt item} plays the same role as $\mathrm{\mathbb{A}_c}.\texttt{item}$ and {\tt klg} stores the knowledge attached on an edge. 

The commonsense knowledge (CSK) we utilize is the social CSK provided by the CSK graph ATOMIC-2020 \cite{ATOMIC2020}. Hwang et al., \shortcite{ATOMIC2020} train a generating model called COMET \cite{COMET} on the CSK graph so that texts out of the scope of the CSK graph can be freely inferred. ATMOIC-2020 contains nine social-interaction relations while not all relations can be fit into our condition. According to the direction of inference and relevance, we pick four relations: \texttt{xEffect}, \texttt{xReact}, \texttt{oEffect}, and \texttt{oReact}. \texttt{x(o)Effect} reasons the effect on the speaker-self (other speakers) after an utterance, and \texttt{x(o)React} describes how the speaker-self (other speakers) feels after an utterance. We exemplify these relations in Appendix A. Given an utterance $u_i$ and one of the relations, COMET takes them as the input and generates knowledge with the beam search of 5. We denote the generation as $\texttt{CT}(u_i, \texttt{xE})$, where \texttt{CT} and \texttt{xE} are the abbreviations of COMET and \texttt{xEffect} respectively. 

As COMET does not guarantee to generate emotion-related knowledge, it is necessary to pick up the knowledge related to the emotion of the targeted utterance. Furthermore, knowledge passing should align with speakers' interactions. Therefore, we propose a knowledge selecting strategy, which is sentiment realized and fits the condition of conversations. 

\begin{algorithm}[tb]
\caption{Construction of Knowledge adjacency matrix}
\label{alg:algorithm}
\textbf{Input}: Dialog $[u_1,...,u_N]$, emotion $[e_1,...,e_N]$, speaker $[p_1,...,p_N]$, COMET \texttt{CT}, window $w_k$, func $split$($\cdot$), $e2s$($\cdot$).\\
\textbf{Output}: Knowledge passing adjacency matrix $\mathrm{\mathbb{A}_k}$. 

\begin{algorithmic}[1] 
\STATE $\forall i,j\in [1,...,N]$, initializing $\mathrm{\mathbb{A}_k}$ with $\mathrm{\mathbb{A}_k}[i,j].\mathrm{item}=0$, $\mathrm{\mathbb{A}_k}[i,j].\mathrm{klg}=$``none''. ($i$: target, $j$: source)
\FOR{$j$ in $[1, N]$}
\STATE $\mathrm{posK_{x}}$, $\mathrm{negK_{x}}$, $\mathrm{neuK_{x}}$ = $split$(\texttt{CT}($u_j$, \texttt{xE/xR}))
\STATE $\mathrm{posK_{o}}$, $\mathrm{negK_{o}}$, $\mathrm{neuK_{o}}$ = $split$(\texttt{CT}($u_j$, \texttt{oE/oR}))
\FOR{$i$ in $[j, \mathrm{min}(j+w_{k}, N)]$}
\IF{$p_i$ == $p_j$}
\STATE $\mathrm{K}$ = \{\texttt{pos}: $\mathrm{posK_{x}}$, \texttt{neg}: $\mathrm{negK_{x}}$, \texttt{neu}: $\mathrm{neuK_{x}}$\}
\ELSE 
\STATE $\mathrm{K}$ = \{\texttt{pos}: $\mathrm{posK_{o}}$, \texttt{neg}: $\mathrm{negK_{o}}$, \texttt{neu}: $\mathrm{neuK_{o}}$\}
\ENDIF
\IF{$e2s$($e_i$) $\neq$ \texttt{neu}}
\STATE $\mathrm{\mathbb{A}_k}[i,j].\mathrm{item}=1$
\IF{$e2s$($e_j$) == \texttt{neu}}
\STATE $\mathrm{\mathbb{A}_k}[i,j].\mathrm{klg}=\mathrm{K}[\texttt{neu}]+[\mathrm{sep}]+\mathrm{K}[e2s(e_i)]$
\ELSE
\STATE $\mathrm{\mathbb{A}_k}[i,j].\mathrm{klg}=\mathrm{K}[e2s(e_i)]$
\ENDIF
\ENDIF
\ENDFOR
\ENDFOR
\STATE \textbf{return} $\mathrm{\mathbb{A}_k}$
\end{algorithmic}
\end{algorithm}

The knowledge selection realizes two factors:

\textbf{Speaker realization}: If $p_i$ of the target $u_i$ equals to $p_j$ of the source $u_j$, knowledge of \texttt{xEffect} and \texttt{xReact} is chosen, otherwise knowledge of \texttt{oEffect} and \texttt{oReact} picked. 

\textbf{Sentiment realization}: For the generated set of knowledge, we need to first identify its emotion, and then pick up emotion-related pieces for $u_i$ and $u_j$. As no standard emotion lexicon resource is provided, we replace the emotion with the sentimental polarity. We utilize sentiwordnet to compute the sentimental polarity of every piece of knowledge. As sentiwordnet rates every word with scores of \texttt{neutral}, \texttt{negative}, and \texttt{positive}, we compute the sentimental score of a piece of knowledge by:
\begin{equation}
    r_{s} = \left\{\begin{matrix}
r_{pos}-r_{neg}, & \left| r_{pos}-r_{neg}\right|> r_{neu} \\
0, & \mathrm{else} \\
\end{matrix}\right.
\end{equation}
where $r_{pos}$ is the average positive score of all words in the knowledge, and the knowledge is regarded as positiveness when $r_s>0$. According to the $r_s$, we split knowledge into three sets and concatenate the knowledge in a set with the token $[\mathrm{sep}]$ (denoted as a function $split(\cdot)$). If a set is empty, the text ``none'' will be assigned to it. After splitting knowledge, we further map the emotion of $u_i$ and $u_j$ into the sentiment (denoted as a function $e2s(\cdot)$). The emotion \texttt{happiness} is mapped to \texttt{positive}, \texttt{neutral} to \texttt{neutral}, and other emotions in the dataset to \texttt{negative}. With the mapped sentiment, we can select the related knowledge for utterances. 

The knowledge selection is processed simultaneously along the construction of $\mathrm{\mathbb{A}_{k}}$, which is illustrated in Algorithm~\ref{alg:algorithm}. In line 14 of Algorithm~\ref{alg:algorithm}, we add the neutral knowledge, because we think that it can bring some reasoning information to help understand causes. 

\subsubsection{Knowledge Enhanced DAG Networks}

As the original DAG network considers no knowledge, we modify it and propose the Knowledge Enhanced DAG network to better process the KEC graph. 

We first encode the knowledge into representations by using the shared RoBERTa of the utterance encoder and the same operation of encoding utterance. For an item of knowledge $\mathrm{\mathbb{A}_{k}}[i,j].\texttt{klg}$, it is encoded as $k_{i,j}\in \mathbb{R}^{d_u}$. 

For a targeted utterance $u_i$, edge weights from neighboring nodes $\mathrm{v}_j$ in the $l^{th}$ layer can be computed as:
\begin{equation}
    \alpha_{i,j}=\mathrm{Softmax}_{j\in\mathcal{N}_{i}}(\mathrm{W}_{\mathrm{w}}^{l}[h_{i}^{l-1}||(h_{j}^{l}+\mathrm{W}_{\mathrm{k}}^{l}k_{i,j})]),
\end{equation}
where $h_{i}^{l-1}$ is the node representation of the $(l-1)^{th}$ layer, $h_{j}^{l}$ is the neighboring node representation that has been processed in the $l^{th}$ layer, $\mathrm{W}_{\mathrm{w}}^{l}\in \mathbb{R}^{1\times2d_u}$ and $\mathrm{W}_{\mathrm{k}}^{l}\in \mathbb{R}^{d_u\times d_u}$ are trainable parameters. For $h_i^0\in\mathbb{R}^{d_u}$, we initialize it by $h_i^0=\mathrm{Linear}([s_{i}||eemb_{e_i}])$, where $eemb_{e_i}$ is the trainable embedding of the emotion $e_i$ and is initialized by the weight vector in the MLM head of RoBERTa. Furthermore, the contextual message and knowledge message are aggregated from neighboring nodes according to the edge weight $\alpha_{i,j}$:
\begin{eqnarray}
    msg_i= \sum_{j\in\mathcal{N}_{i}}{\alpha_{i,j}\mathrm{W}_{\mathbb{A}_{\mathrm{c}[i,j].\texttt{rel}}}^{l}h_{j}^{l}}, \\
    nlg_i = \sum_{j\in\mathcal{N}_{i}}{\alpha_{i,j}\mathrm{W}_{\mathrm{k}}^{l}k_{i,j}},
\end{eqnarray}
where $\mathrm{W}_{\mathbb{A}_{\mathrm{c}[i,j].\texttt{rel}}}\in(\mathrm{W}_{\mathrm{SD}}^{l},\mathrm{W}_{\mathrm{ID}}^{l})$ with the size of $\mathbb{R}^{d_u\times d_u}$ is a trainable parameter. 

\begin{table}[]
\centering
\scalebox{0.8}{
\begin{tabular}{cc|c|c|c}
\toprule
                                                  &          & Train & Dev  & Test  \\ \hline\hline
\multicolumn{1}{c|}{\multirow{2}{*}{\begin{tabular}[c]{@{}c@{}}Num. of \\ Causal Pair\end{tabular}}} & Positive & 7027  & 328  & 1767  \\ \cline{2-5}
\multicolumn{1}{c|}{}                             & Negative & 45392 & 2842 & 14052 \\ \hline
\multicolumn{2}{c|}{Num. of Dialgoue}                           & 834   & 47   & 225   \\ \hline
\multicolumn{2}{c|}{Num. of Utterance}                          & 8206  & 493  & 2405  \\ \hline
\multicolumn{2}{c|}{Avg. Len. of Utterance}                     & 14    & 16   & 15    \\ \bottomrule
\end{tabular}
}
\caption{Statistics of RECCON-DD. ``Positive'' means the true causal pair. }
\vspace{-0.2cm}
\label{dataset}
\end{table}

To capture the nodal and contextual information, DAG networks utilize two GRUs: \textit{nodal unit} ($\mathrm{GRU}_{n}$) and \textit{contextual unit} ($\mathrm{GRU}_{c}$). The information is obtained by:
\begin{eqnarray}
    nod_{i}=\mathrm{GRU}_{n}(h_{i}^{l-1}, msg_{i}), \\
    cxt_{i}=\mathrm{GRU}_{c}(msg_{i}, h_{i}^{l-1}). 
\end{eqnarray}

To further consider the effect of knowledge from the context and the utterance itself, we employ two additional GRUs: \textit{contextual knowledge unit} ($\mathrm{GRU}_{k}$) and \textit{self-loop knowledge unit} ($\mathrm{GRU}_{s}$). Therefore, knowledge-related information can be obtained by:
\begin{eqnarray}
    ckg_{i}=\mathrm{GRU}_{k}(nlg_i,  h_{i}^{l-1}), \\
    skg_{i}=\mathrm{GRU}_{s}(k_{i,i},  h_{i}^{l-1}). 
\end{eqnarray}

Finally, the node representation of utterance $u_i$ in the $l^{th}$ layer is updated by summing the four types of information, i.e. $h_{i}^{l}=nod_i + cxt_i + ckg_i + skg_i$. 
\subsection{Cause Predictor}
After the encoding of two-level encoders, the final utterance representation is computed by $h_i=||_{l=0}^{L}h_{i}^{l}$ \cite{DAG}. Whether $u_j$ is the cause of $u_i$ is then computed by:
\begin{equation}
    p_{i,j}=sigmoid(\mathrm{MLP}([h_i||h_j])), 
\end{equation}
where MLP maps the concatenation of $h_i$ and $h_j$ from the dimension of ${2d_u}$ to 1. 

We use the cross entropy loss to train our model, and the loss on a conversation is formulated as:
\begin{equation}
    \mathcal{L}=\sum_{i\leq N}\sum_{j\leq i}(y_{i,j}\cdot \mathrm{log}p_{i,j}+(1-y_{i,j})\cdot(1-p_{i,j})), 
\end{equation}
where $y_{i,j}\in \{0, 1\}$ is the cause label of the pair $(u_i, u_j)$. 

\section{Experimental Setup}

\subsection{Dataset}

We conduct experiments on the dataset call RECCON-DD \cite{RECCON}, which is collected from the popular dyadic dialogue dataset DailyDialog (DD) \cite{DailyDialog}. Preprocessing the original RECCON-DD, we only consider causes in the dialogue history for every utterance and remove the duplicated causal pairs. Statistics of the processed RECCON-DD are presented in Tab.~\ref{dataset}. 

\begin{table}[]
\centering
\scalebox{0.8}{
\begin{tabular}{@{}c|l|ccc@{}}
\toprule
                   &\multicolumn{1}{l|}{Model}     & Neg. F1     & Pos. F1     & \multicolumn{1}{c}{macro F1}    \\ \midrule\midrule
\multirow{3}{*}{1} &\multicolumn{1}{l|}{ECPE-2D$^\triangle$}    & 94.96       & 55.50       & 75.23       \\
                   &\multicolumn{1}{l|}{ECPE-MLL$^\triangle$}   & 94.68       & 48.48       & 71.59       \\
                   &\multicolumn{1}{l|}{RankCP$^\triangle$}     & 97.30       & 33.00       & 65.15       \\ \midrule
\multirow{3}{*}{2} &\multicolumn{1}{l|}{KAG}        & 94.49$_{(0.22)}$ & 55.52$_{(2.39)}$ & 75.02$_{(1.13)}$ \\
                   &\multicolumn{1}{l|}{Adapted}    & 95.67$_{(0.24)}$ & 62.47$_{(4.72)}$ & 79.07$_{(2.44)}$ \\
                   &\multicolumn{1}{l|}{SKAIG}      & 95.26$_{(0.12)}$ & 63.15$_{(1.00)}$  & 79.21$_{(0.49)}$ \\ \midrule
\multirow{2}{*}{3} &\multicolumn{1}{l|}{Entail}     & 94.83$_{(0.47)}$ & 58.59$_{(3.78)}$ & 76.66$_{(1.66)}$ \\
                   &\multicolumn{1}{l|}{DAG-ERC}    & 95.33$_{(0.25)}$ & 63.56$_{(2.10)}$ & 79.44$_{(1.16)}$ \\ \midrule
4                  &\multicolumn{1}{l|}{KEC (ours)} & 95.74$_{(0.05)}$ & \textbf{66.76$^{*}_{(0.33)}$} & \textbf{81.25$^{*}_{(0.17)}$} \\ \bottomrule
\end{tabular}
}
\caption{Results of all models on RECCON-DD. $^*$ denotes that our method is significant against the best baseline DAG-ERC (p-value$<$0.05) with the paired T-test. $^\triangle$ denotes the results referred from Poria et al.~[2021]. }
\vspace{-0.05cm}
\label{main}
\end{table}

\subsection{Compared Methods}

For the reason that C${_2}$E${_2}$ is a new task proposed by Poria et al. \shortcite{RECCON}, there is only one baseline \cite{RECCON}. Therefore, we additionally compare our model with other SOTA methods from conversation and emotion cause related tasks. 

\textbf{Entail} is the baseline of C${_2}$E${_2}$. It utilizes RoBERTa to process the concatenation of utterances and their emotion and history context. The concatenation is formed by ``[cls] emotion [sep] $u_i$ [sep] $u_j$ [sep] history context [sep]''. \textbf{DAG-ERC} \cite{DAG} is the SOTA method of Conversational Emotion Recognition (ERC). Our method is based on it. \textbf{SKAIG} \cite{SKAIG} is another SOTA method of ERC which utilize CSK to help the prediction. It passes knowledge through edges using graph transformers \cite{TransformerConv}, while no knowledge selecting strategy is applied. For the fair comparison, we utilize the knowledge \texttt{x/oEffect} and \texttt{x/oReact} for SKAIG. \textbf{KAG} \cite{KAG} is the SOTA method of Emotion Cause Extraction (ECE) using CSK to overcome the position bias. It connects emotional utterances with a contextual utterance if a path is matched in ConceptNet. It uses relational GCNs to process their proposed graph. Knowledge used in KAG only participates in the computation of edge weights. As other methods add no position embeddings, we do not add position embeddings in KAG. \textbf{Adapted} \cite{AdaptedKnow} is another SOTA method of ECE. It directly concatenates 
CSK (\texttt{xReact} and \texttt{oReact}) after the utterance and utilizes RoBERTa to make prediction. For baselines from Emotion Cause Pair Extraction (ECPE), following and referring the results from Poria et al.~\shortcite{RECCON}, we choose \textbf{ECPE-2D} \cite{ECPE_2D}, \textbf{ECPE-MLL} \cite{SlidingWindow}, and \textbf{RankCP} \cite{Rank_Emotion}. To consider emotion information, in the re-implementation, all graph-related baselines utilize emotion embedding. 

\begin{table}[]
    \centering
    \scalebox{0.8}{
    \begin{tabular}{@{}l|c|c@{}}
\toprule
~Method                                                                                   & Pos. F1             & ~~macro F1~~~            \\ \hline\hline
~KEC                                                                                      & ~\textbf{66.76$_{(0.33)}$}    & ~\textbf{81.25$_{(0.17)}$}    \\ \hline
\multirow{2}{*}{\begin{tabular}[c]{@{}l@{}}~$-$ \textit{contextual}\\ ~~~~~\textit{knowledge unit}~~\end{tabular}} & ~64.93$_{(0.69)}$    & ~80.25$_{(0.37)}$    \\
                                                                                         & \multicolumn{1}{l|}{$-$1.83$_{(\downarrow)}$}  & \multicolumn{1}{l}{$-$1.00$_{(\downarrow)}$}  \\ \hline
\multirow{2}{*}{\begin{tabular}[c]{@{}l@{}}~$-$ \textit{self-loop}\\ ~~~~~\textit{knowledge unit}\end{tabular}}  & ~65.00$_{(0.92)}$    & ~80.21$_{(0.54)}$    \\
                                                                                         & \multicolumn{1}{l|}{$-$1.76$_{(\downarrow)}$}  & \multicolumn{1}{l}{$-$1.04$_{(\downarrow)}$}  \\ \hline
\multirow{2}{*}{\begin{tabular}[c]{@{}l@{}}~$-$ neutral\\ ~~~~~knowledege\end{tabular}}        & ~65.80$_{(0.96)}$    & ~80.67$_{(0.52)}$    \\
                                                                                         & \multicolumn{1}{l|}{$-$0.96$_{(\downarrow)}$}  & \multicolumn{1}{l}{$-$0.58$_{(\downarrow)}$}  \\ \hline
\multirow{2}{*}{\begin{tabular}[c]{@{}l@{}}~$-$ emotion \\ ~~~~~embs\end{tabular}}             & ~63.19$_{(0.25)}$                   & ~79.35$_{(0.17)}$                   \\
                                                                                         & \multicolumn{1}{l|}{$-$3.56$_{(\downarrow)}$}  & \multicolumn{1}{l}{$-$1.90$_{(\downarrow)}$}  \\ \hline
\multirow{2}{*}{\begin{tabular}[c]{@{}l@{}}~$-$ emoition embs \&\\ ~~~~~CSK\end{tabular}}      & ~46.66$_{(0.76)}$    & ~69.90$_{(0.48)}$    \\
                                                                                         & \multicolumn{1}{l|}{$-$20.1$_{(\downarrow)}$} & \multicolumn{1}{l}{$-$11.4$_{(\downarrow)}$} \\ \bottomrule
\end{tabular}
    }
    \caption{Ablation Study}
    \vspace{-0.2cm}
    \label{ablation}
\end{table}

\subsection{Implementation}

All compared methods and our model use base-size pretrained models (e.g., RoBERTa-base) from \textit{huggingface} \textit{Transformers}. We train our model using the AdamW optimizer with 40 epochs, learning rate of 3e-5, L2 regularization of 1e-4. The batch is set to 4 and we step the gradient with the accumulating step of 2. For Conversation Encoder, we set the dimension of utterance to 300, the hidden size of GRUs to 300, the number of layers of DAG networks to 5. For emotion embeddings, we utilize a linear unit mapping the dimension of 768 to 200. For Cause Predictor, dimensions of MLP is set to [600, 300, 300, 1] and the dropout rate is set to 0.1. For the window size of knowledge and that of context, we find that our model achieves better performance when they are the same, and we set them to 2.  Following Poria et al.~\shortcite{RECCON}, we report the F1 scores of both negative and positive causal pairs and the macro F1 score of them. Our experiments are conducted using 5 random seeds.

\section{Results and Discussions}

\begin{figure*}
    \centering
    \scalebox{0.385}{\includegraphics{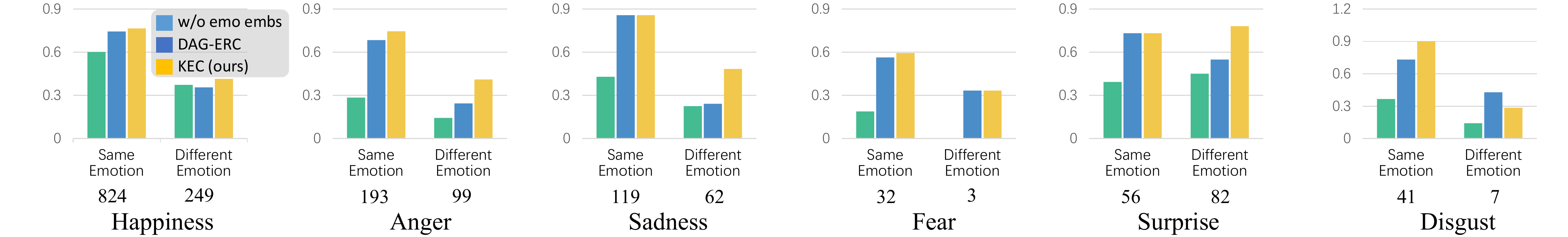}}
    \caption{``Same Emotion'' reports the recall of causal pairs whose causal utterances are with the same emotion as the targeted utterance. ``Different Emotion'' refers to different emotions from the targeted utterance. The number of a type of pairs is presented below the x-axis. Green bars denote DAG-ERC without emotion embeddings, blue bars for DAG-ERC with emotion embeddings, and orange bars for KEC. }
    \vspace{-0.1cm}
    \label{emotion}
\end{figure*}

\subsection{Main Results}

The results of our method and baselines are listed in Tab.~\ref{main}. We split the table into 4 rows: the 1st row for models of ECPE, the 2nd row for models using CSK, the 3rd row for models with no CSK, and the 4th row for our method. 

From the 1st row, models of ECPE are not competitive to models in other rows. We think the reason is that the setting of ECPE is not suitable for C$_2$E$_2$. Compared with models in the 2nd row, our method can achieve better results. For KAG, we think the undesired performance can be attributed to that it is designed to resist position bias for news articles. Conversations in RECCON tends to be short and requires more sophisticated modeling than articles. In addition, knowledge is only used to compute edge weight in KAG, which does not propagate through edges like ours. For ``Adapted'', knowledge is directly concatenated after the utterance, and is not explicitly passed between utterances. For SKAIG, knowledge is also not selected, which leads to sentiment-nonaligned knowledge passed through the graph. On the contrary, our KEC selects proper knowledge to align with emotions, and explicitly models and propagates the knowledge in graphs. The comparison indicates that our method is more suitable for utilizing knowledge in C$_2$E$_2$. 

As the baseline of RECCON, ``Entail'' is not comparable to the graph-based SKAIG and DAG-ERC, which demonstrates that explicit modeling of utterance interactions is critical for C$_2$E$_2$. DAG-ERC is the SOTA baseline for ERC with no knowledge involved. By introducing CSK and the knowledge selection, our method can achieve significant improvements against DAG-ERC. Therefore, it is critical to utilize proper knowledge for cause detection. To verify our knowledge enhanced DAG networks, we construct a variant that knowledge $\mathrm{W}_{\mathrm{k}}^{l}k_{j,i}$ is directly added on $h_j^l$ in eq.~4 and no knowledge-related GRUs are involved. The macro F1 and Pos. F1 achieved by the variant are 80.38 and 65.24. This demonstrates that, instead of being simply added in contextual messages, knowledge prefers to be exclusively modelled by gated units in our knowledge enhanced DAG networks. 

\begin{figure}
    \centering
    \vspace{-0.3cm}
    \scalebox{0.43}{\includegraphics{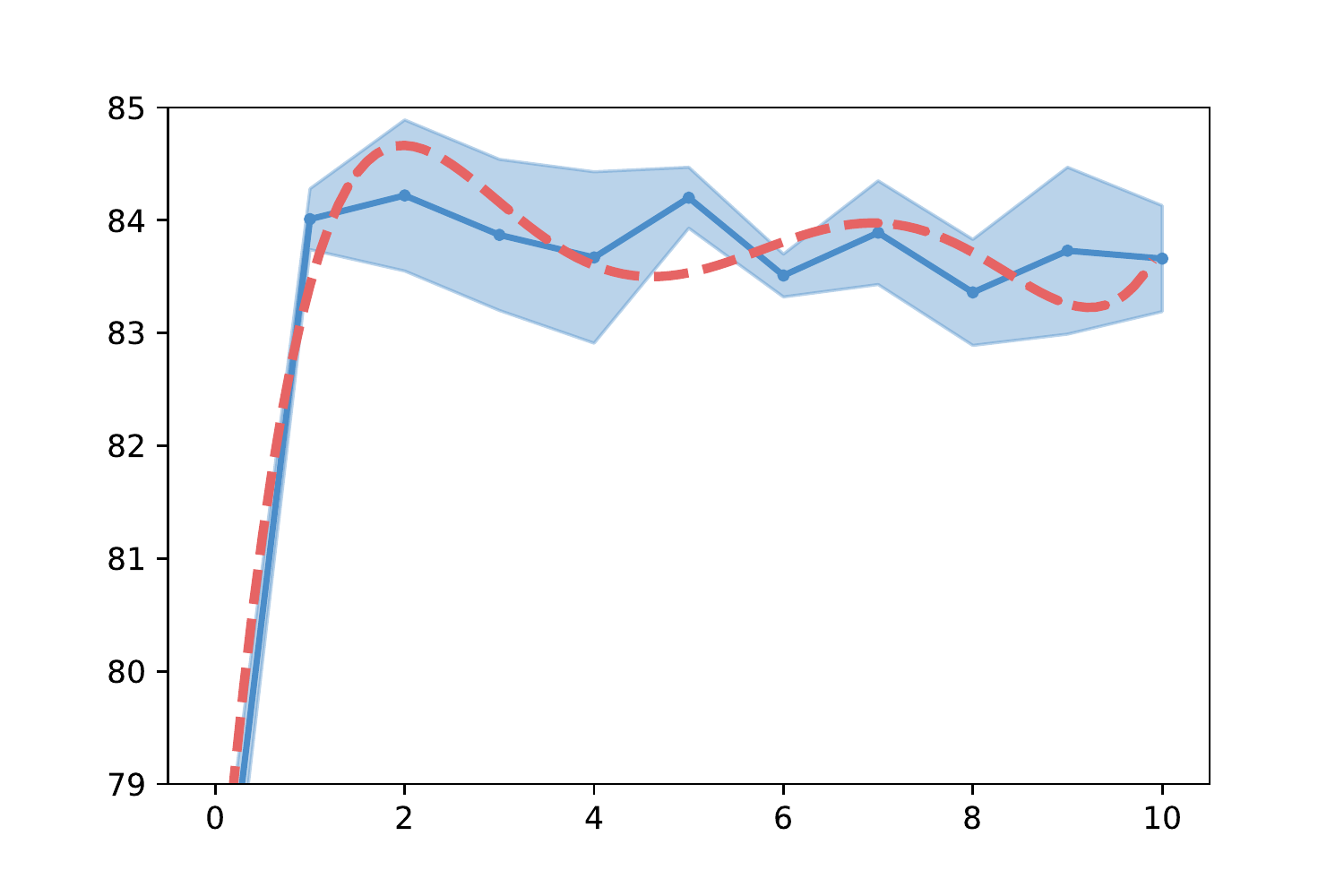}}
    \vspace{-0.2cm}
    \caption{The effect of the window size. The y-axis denotes the macro F1 on the development set. To better fit the trend, we draw the trend line of fifth-order polynomial as the red dotted line. }
    \vspace{-0.2cm}
    \label{window}
\end{figure}

\subsection{Ablation Study}

To study the effect of different modules equipped in our model, we remove \textit{contextual knowledge unit} $\mathrm{GRU}_k$, \textit{self-loop knowledge unit} $\mathrm{GRU}_s$, neutral knowledge added when source utterances are neutral in line 14 of Algorithm 1, and emotion embedding respectively. The results are in Tab.~\ref{ablation}. 

The decreased performance by removing $\mathrm{GRU}_k$ shows the necessity of controlling knowledge from the context. As for removing $\mathrm{GRU}_s$ affects the performance, it indicates that self-knowledge enriching for targeted utterances is critical because 37$\%$ causal utterances are the targeted utterances themselves in RECCON. By taking off neutral knowledge, the performance shows decline. This manifests that neutral knowledge can help the model to better detect causes as it can provide some reasoning information. 

When removing emotion embedding, the performance is still competitive to DAG-ERC who is equipped with emotion embedding. This can be attributed to that our knowledge selecting strategy can pick up sentiment related knowledge for the model. As emotion embedding and knowledge are all removed, the model becomes DAG-ERC without emotion information. The performance drops dramatically, which verify the claim that emotion information are important to C$_2$E$_2$. As for ``Entail'' without emotion, the performance is presented in Appendix C. In the next subsection, we will further show what causes are preferred by the emotion information. 


\subsection{Performance on Cause Pairs}

There is a question about whether our model can promote the detection of causal utterances with emotions different from that of the targeted utterance. We therefore present the recalls of causal pairs whose utterances are with the same emotion (SE pairs) and different emotions (DE pairs) in Fig.~\ref{emotion} for every targeted emotion. From the figure, we can see that DAG-ERC with emotion embedding can dramatically boost the detection of SE pairs while limited improvement is achieved on DE pairs. Specifically, emotion information increases the recall on all SE pairs by about 23$\%$ while only about 3$\%$ on DE pairs. By introducing CSK, our model, compared with DAG-ERC with emotion embedding, further increases the recall by 3$\%$ on SE pairs and by 12$\%$ on DE pairs (especially boosting the detection of neutral causal utterances by 12$\%$). This manifests that our model can promote the detection of DE pairs, especially on pairs whose targeted utterances are with negative emotions like \texttt{Anger}, \texttt{Sadness} in Fig.~\ref{emotion}. 

\subsection{Effect of Window Size}

To show the effect of the window size, we illustrate the trend of macro F1 scores with the window size increasing in Fig.~\ref{window}. The trend line shows that our model achieves the best performance with a small window. As the window size keeps increasing, the performance will drop and fluctuate. The reason for this can be attributed to that over 80$\%$ causal utterances are located near the targeted utterance in no more than 2 steps. It can be future work to solve the problem that models bias to neighboring utterances in C$_2$E$_2$. 

\section{Conclusion}

In this paper, we study \textit{Conversational Causal Emotion Entailment}. We find that adding emotion information into graphs cannot effectively detect causal utterances with different emotions from the targeted utterance due to models' limited causal reasoning ability. To alleviate this, we introduce social commonsense knowledge and propose a graph-based structure to properly utilize knowledge. Specifically, we propose a Knowledge Enhanced Conversation graph (KEC) with a designed knowledge selecting strategy. To process KEC, we construct Knowledge Enhanced Directed Acyclic Graph networks. Our method outperforms baselines and detects more causes with different emotions from the targeted utterance.  

\section*{Acknowledgments}

This work was supported by National Natural Science Foundation of China (No. 61976207, No. 61906187). 

\bibliographystyle{named}
\bibliography{ijcai22}

\end{document}